\documentclass[11pt]{article}
\usepackage[margin=0.91in]{geometry} 
\usepackage{kpfonts}
\usepackage[colorlinks,
citecolor=blue
]{hyperref}  
\usepackage{url}
\usepackage{bm}
\usepackage{microtype}
\usepackage{graphicx}
\usepackage{subfigure}  % support sub-figure
\usepackage{booktabs} % for professional tables 

\usepackage{hyperref}
\usepackage{algorithm, algorithmic}

\usepackage{natbib}

\usepackage{multirow}
\usepackage{colortbl}
\usepackage{booktabs,multirow,array,caption}
\usepackage{pbox}
\usepackage{pifont}% http://ctan.org/pkg/pifont
\usepackage{tikz}
\usepackage{xcolor}
\usepackage{graphicx}

\DeclareSymbolFont{extraup}{U}{zavm}{m}{n}
\DeclareMathSymbol{\varheartsuit}{\mathalpha}{extraup}{86}
\DeclareMathSymbol{\vardiamondsuit}{\mathalpha}{extraup}{87} 

\usepackage{amsfonts}
\usepackage{xcolor}
\let\proof\relax
\let\endproof\relax
\usepackage{amsthm}
\usepackage{amsmath}
\usepackage{amssymb}
\usepackage{soul}
\usepackage{comment}
\usepackage{enumitem}
\usepackage{bbm}
\usepackage{graphicx}
\usepackage{float}
\usepackage{mathtools}

\theoremstyle{remark}

\renewcommand{\epsilon}{\varepsilon}

\usepackage{array}
\newcolumntype{M}[1]{>{\centering\arraybackslash}m{#1}}

\usepackage{tablefootnote}
\usepackage[symbol]{footmisc}

\title{VMAgent: Scheduling Simulator for Reinforcement Learning}

\makeatletter
\def\@fnsymbol#1{\ensuremath{\ifcase#1\or  \natural \or \dagger\or * \or \ddagger\or
		\mathsection\or \mathparagraph\or \|\or **\or \dagger\dagger
		\or \ddagger\ddagger \else\@ctrerr\fi}}
\makeatother

\author{\normalsize Junjie Sheng\textsuperscript{\dag}\thanks{School of Computer Science and Technology, East China Normal University, Shanghai, China. Email addresses: \{jarvis@stu, slcai@stu, 51215901007@stu, whli@stu,  yunhua@stu, bjin@cs, xfwang@cs\}.ecnu.edu.cn.} \and
Shengliang Cai\textsuperscript{\dag} \and 
Haochuan Cui\textsuperscript{\dag} \and
Wenhao Li\textsuperscript{\dag} \and  
Yun Hua\textsuperscript{\dag} \and
Bo Jin\textsuperscript{\dag} \and 
Wenli Zhou\textsuperscript{\ddag}\thanks{Algorithm Innovation Lab, Cloud BU, Huawei Technologies Co., Ltd. Email addresses: \{zhouwenli, huyiqiu1\}@huawei.com} \and
Yiqiu Hu\textsuperscript{\ddag} \and
Lei Zhu\textsuperscript{$\natural$}\thanks{Alkaid Lab, Cloud BU, Huawei Technologies Co., Ltd. Email addresses:\{zhulei41, pengqian19\}@huawei.com} \and
Qian Peng\textsuperscript{$\natural$} \and 
Hongyuan Zha\textsuperscript{\S}\thanks{The Chinese University of Hong Kong-Shenzhen. Email address: zhahy@cuhk.edu.cn}\and
Xiangfeng Wang\textsuperscript{\dag}
}

\date{\normalsize }

\begin{document}
 
 \maketitle

\begin{abstract}
A novel simulator called \texttt{VMAgent} is introduced to help RL researchers better explore new methods, especially for virtual machine scheduling.
\texttt{VMAgent} is inspired by practical virtual machine (VM) scheduling tasks and provides an efficient simulation platform that can reflect the real situations of cloud computing.
Three scenarios (fading, recovering, and expansion) are concluded from practical cloud computing and corresponds to many reinforcement learning challenges (high dimensional state and action spaces, high non-stationarity, and life-long demand).
\texttt{VMAgent} provides flexible configurations for RL researchers to design their customized scheduling environments considering different problem features.
From the VM scheduling perspective, \texttt{VMAgent} also helps to explore better learning-based scheduling solutions.
\end{abstract}

\section{Introduction}
\label{introduction}
Reinforcement learning (RL) has shown competitive performance in games~\citep{go,dota} and robotics simulators~\citep{ppo,james20163d}.
Recently, it has received extreme attention to solving mathematical optimization problems (e.g., linear programming, mixed-integer programming, combinatorial optimization) with RL methods~\citep{nair2020solving}.
Scheduling is one of the typical mathematical optimization problems. It widely exists in real-world applications~\citep{shyalika2020reinforcement}, like cloud computing, transportation, manufacturing, etc.
Especially in cloud computing, virtual machine scheduling is the core of Infrastructure as a Service (IaaS)~\citep{liu2016survey, azure}.
Many classical combinatorial optimization methods were used to solve offline VM scheduling problem~\citep{WOLKE201583}. However, practical scheduling scenarios mainly rely on heuristic methods~\citep{bays1977comparison} due to the online requirement.
Unfortunately, Heuristic methods heavily depend on expert knowledge and might get stuck in sub-optimal solutions.
The RL-based method has great potential to solve VM scheduling problems, and primary advantages have been shown in \citet{hw}.
To explore further with RL, an efficient and realistic VM scheduling simulator is necessary to be proposed.

In this paper, we propose a novel VM scheduling simulator called \texttt{VMAgent}, based on real data from Huawei Cloud's actual operation scenarios.
\texttt{VMAgent} aims to simulate the scheduling process of virtual machine requests (allocating and releasing CPU and memory resources) on multiple servers.
It builds virtual machine scheduling scenarios based on practical system architecture, including fading, recovering, and expansion.
The fading scenario allows only requests allocation, while the recovering scenario allows both allocating and releasing VM resources.
Different from them, the expansion scenario allows adding servers to avoid termination, which is common in the growing public cloud.
Our \texttt{VMAgent} provides flexible configurations to define these scenarios.
\texttt{VMAgent} provides an efficient simulation platform, which can reflect the real situations of cloud computing.
Furthermore, \texttt{VMAgent} provides simple but powerful visualizations for deeply understanding and effectively comparing VM scheduling algorithms.
From the perspective of VM scheduling, few companies open-sourced their cloud data~\citep{azure}.
The few open-sourced data often contain redundant information, and critical information is often transformed into less meaningful float numbers.
Thus our \texttt{VMAgent} is critical for exploring novel methods on VM scheduling, especially for RL methods.

Given the excellent simulating performance of \texttt{VMAgent}, it is an ideal platform for RL-based VM scheduling algorithms. 
Besides the flexible and efficient advantage mentioned above, it can cover both VM scheduling and RL challenges.
\texttt{VMAgent} can accommodate a large number of servers, which leads to the high dimensional state and action spaces.
It can also set the VM request sequences dynamically similar to real clouds (contain the various size of allocating and release requests), which brings in the high non-stationarity.
\texttt{VMAgent} has an expansion mechanism for continuous execution, which stands for the life-long demand.
These challenges are also three key aspects to apply RL to real-world problems.
From the RL perspective, although there are many simulators for RL, such as Atari~\citep{mnih2015human}, MuJoCo~\citep{ppo} and Dota 2~\citep{dota}, there are still no simulators for cloud computing.
%From the RL perspective, although there are many simulators for RL, such as Atari\citep{mnih2015human}, Mujoco\citep{ppo} and Dota2\citep{dota}, most of them are not flexible enough to the three challenges.
% To the best of our knowledge, 
To sum up, \texttt{VMAgent} is the first virtual machine simulator with real-world data.
It provides a powerful platform to explore the law and scope of the VM scheduling problem and help design and test efficient RL methods. 
Therefore, it contributes to both VM scheduling and RL communities.

\section{The \texttt{VMAgent} Platform}

The \texttt{VMAgent} project\footnote{https://github.com/mail-ecnu/VMAgent} is a reinforcement learning platform for virtual machine scheduling.
\texttt{VMAgent} can build virtual machine scheduling scenarios based on
practical system architecture and open-source VM scheduling dataset, \texttt{Huawei-East-1}\footnote{https://github.com/mail-ecnu/VMAgent/blob/master/vmagent/data/dataset.csv},  from {\em{Huawei Cloud}}.
In addition, it also provides flexible configurations to define a variety of scheduling scenarios that correspond to the aforementioned challenges in applying RL to real-world problems.
Finally, \texttt{VMAgent} can provide simple but powerful visualization schemes for deeply understanding and effectively comparing VM scheduling algorithms.

\begin{figure}[th!]
	\centering
	\includegraphics[width=0.7\textwidth]{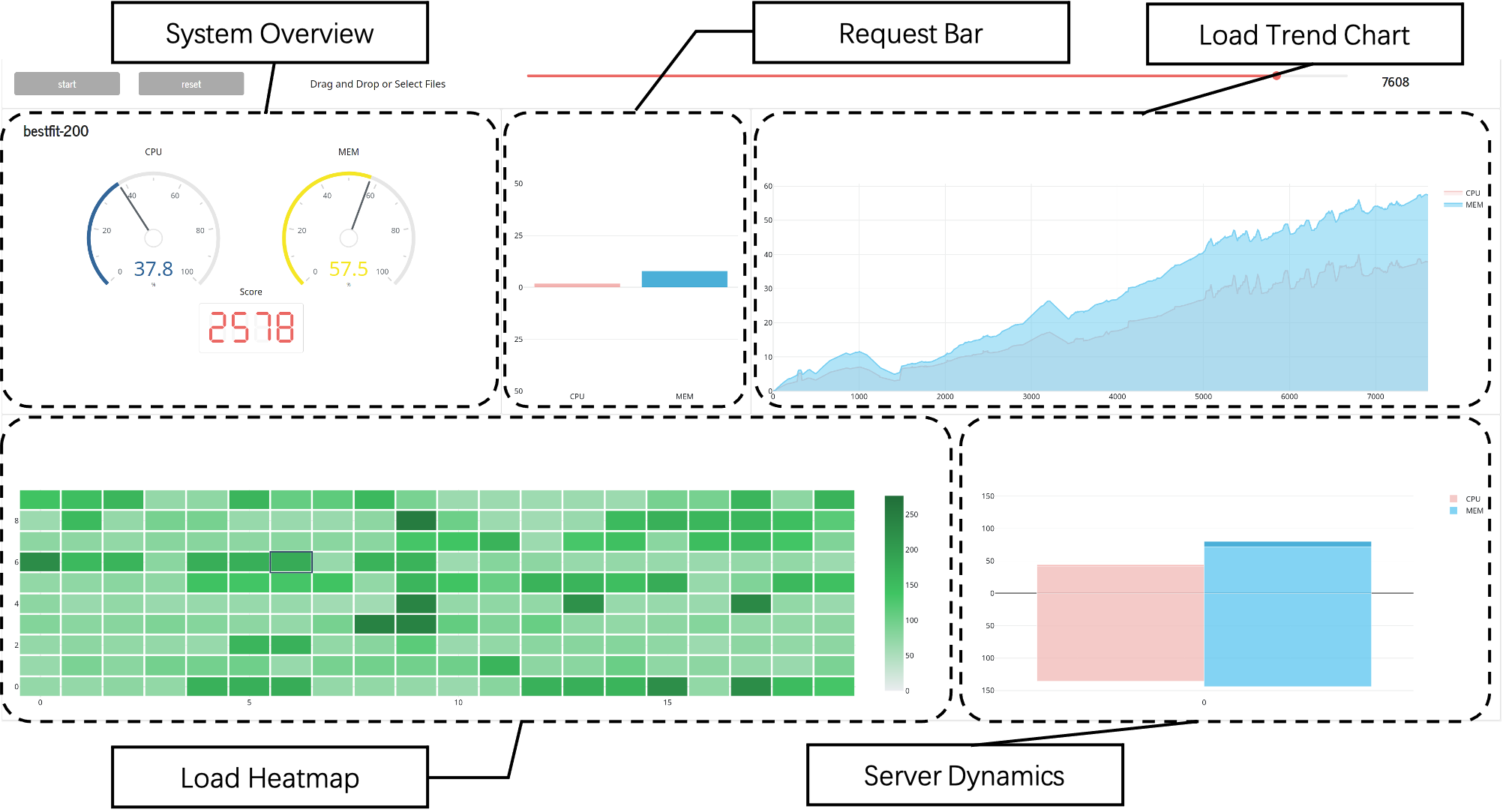}
    \caption{Platform visualization.}
    \label{fig: platform}
    
\end{figure}

\subsection{Scheduling Fundamentals}
The \texttt{Huawei-East-1} dataset was collected in the east china region of Huawei Cloud for one month.
It includes $241743$ requests and $15$ types virtual machines.
In our scheduling environment, we define the agent's observation as the combination of cluster status and the current request information.
The cluster status includes the number of servers within the cluster, the resource occupancy of each server, and servers' specific architecture (double NUMA~\citep{lameter2013numa}).
The current request information can be categorized into two types: 1) allocation request, which includes the feature of resources, such as the amount of needed CPU and memory; 2) release request, which only includes the VM ID.
If an allocation request is encountered, the scheduler needs to select a server (identified by a unique VM ID) with correspondingly occupied resources
to place the request.
If a release request is encountered, the cluster will release the resources occupied by the specified VM.
%, which does not require the scheduler to choose.

\subsection{Configuration Language}
We utilize YAML to provide flexible configurations.
The configurations can be categorized into two types: cluster configuration and scene configuration.
The following example shows an expansion scenario which will be described in detail in the following subsection.
\begin{small}
\begin{verbatim}
cluster_args:   env_args: 
    N: 100            allow_release: True
    CPU: 40           growing_threshold: 0.8 
    MEM: 90           growing_nums: 20
    double_numa: True 
\end{verbatim}
\end{small}

\subsection{Baseline Algorithms}
To help researchers better exploit the platform, we implemented two fundamental heuristic methods (first-fit and best-fit~\citep{bays1977comparison}), which are often adopted in current practical scheduling scenarios.
Further, many RL methods (DQN~\citep{mnih2015human}, A2C~\citep{mnih2016asynchronous}, PPO~\citep{ppo}, SAC~\citep{haarnoja2018soft}, Sched-Q~\citep{hw}) are provided as baselines. 
In the future, more baseline algorithms will be added, including some combinatorial optimization methods.
%{\color{red}{operations research methods}}

\begin{figure}[h]
	\centering
	\includegraphics[width=0.7\textwidth]{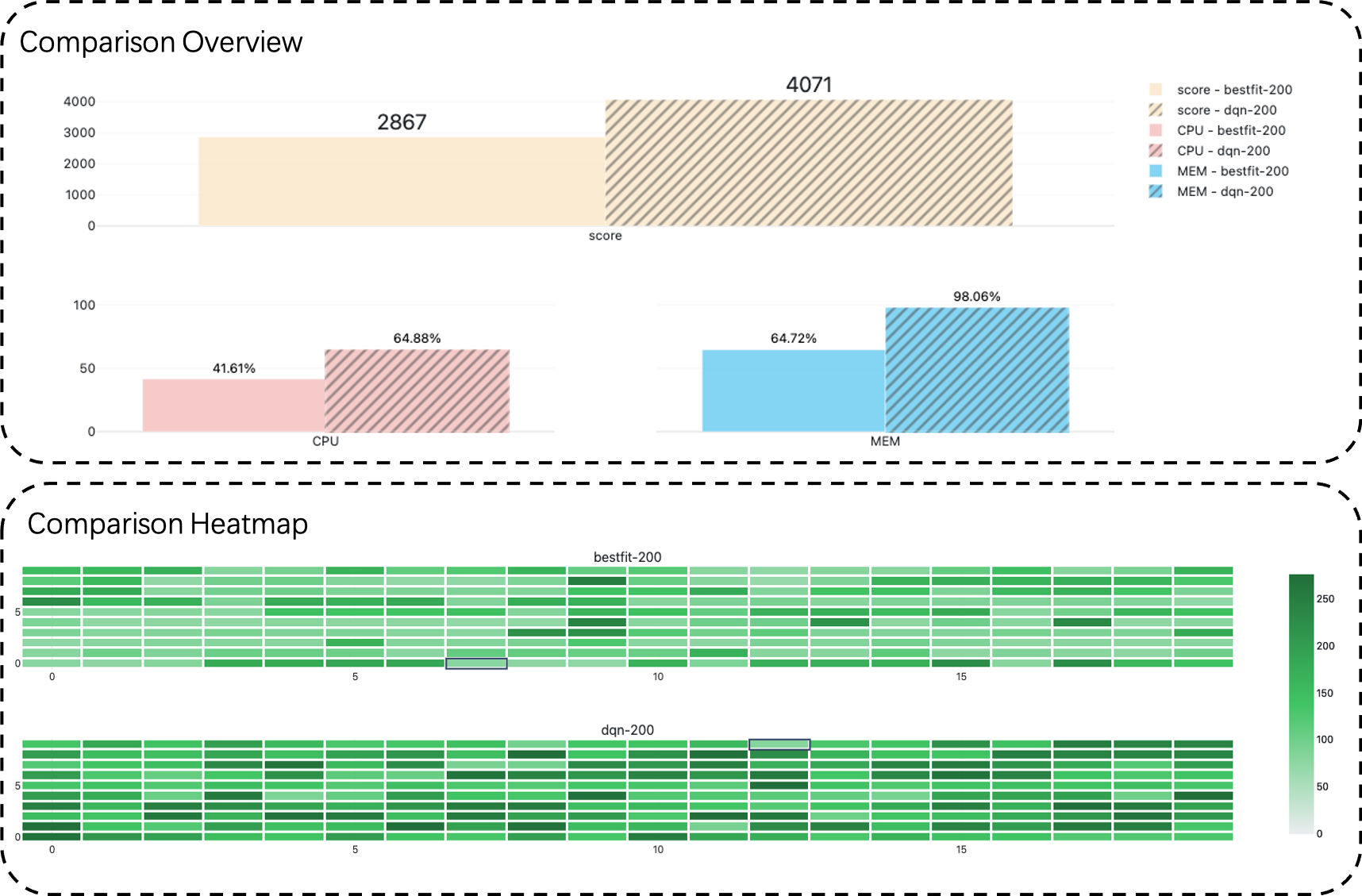}
    \caption{Comparisons between best-fit and DQN.}
    \label{fig: bf2dqn}
\end{figure}

\section{Scenarios and Visualization}
In \texttt{VMAgent}, we provide three scenarios that came from the practical scheduling, i.e., \textit{recovering}, \textit{fading} and \textit{expansion}.
% Each scenario corresponds to one of the aforementioned challenges about applying RL to real-world problems.
%The three scenarios are \textit{fading}, \textit{recovering}, \textit{expansion}.

\noindent{\em{Recovering}}. The recovering scenario considers both allocating and releasing requests, which is common in the public cloud when the resource pool will not be expanded.
The corresponding resources are released from the cluster when the release requests come.
Due to the continual unpredictable release requests, the environment faces high non-stationarity.

\noindent{\em{Fading}}. The fading scenario only allows allocation requests, which is common in the dedicated cloud.
When the number of servers is large, the high-dimension issue comes.

\noindent{\em{Expansion}}. The expansion scenario considers that several servers will be expanded if the remaining resources are lower than a certain threshold, common in the public cloud when the resource pool can be expanded.
The cloud cluster will add several servers before being terminated.
This scenario leads to a life-long VM scheduling problem.

Finally, to deeply understand and compare VM scheduling algorithms (especially with baselines),
we provide the visualization module of our platform as shown in Figrues~\ref{fig: platform} and \ref{fig: bf2dqn}.
This module could visualize the dynamic of cluster status under the corresponding VM scheduling algorithm when continually handling requests (Figrue~\ref{fig: platform}) and the comparison between different algorithms (Figure~\ref{fig: bf2dqn}).

\section{Conclusion}
In this paper, we introduce the \texttt{VMAgent} simulator, to help the RL community investigating critical challenges in applying RL to real-world problems.
\texttt{VMAgent} can also benefit to VM scheduling community, which can be employed to design efficient RL-based VM scheduling methods.
\texttt{VMAgent} contains simple but powerful visualization module for 
understanding and comparing VM scheduling algorithms.

\bibliography{main}
\bibliographystyle{main}
% \EOD

\end{document}